# Analyzing Wearables Dataset to Predict ADLs and Falls: A Pilot Study

Rajbinder Kaur[a*✉] [0000-0002-4996-8915], Rohini Sharma[b][0000-0002-3388-7619]

*rajbinderk54@gmail.com[a], rohini@pu.ac.in[b]*
[a,b] *Department of Computer Science and Applications, Panjab University, Chandigarh(160014), India*

**Abstract**

Healthcare is an important aspect of human life. Use of technologies in healthcare has increased manifolds after the pandemic. Internet of Things based systems and devices proposed in literature can help elders, children and adults facing/experiencing health problems. This paper exhaustively reviews thirty-nine wearable based datasets which can be used for evaluating the system to recognize Activities of Daily Living and Falls. A comparative analysis on the SisFall dataset using five machine learning methods i.e., Logistic Regression, Linear Discriminant Analysis, K-Nearest Neighbor, Decision Tree and Naive Bayes is performed in python. The dataset is modified in two ways, in first all the attributes present in dataset are used as it is and labelled in binary form. In second, magnitude of three axes(x,y,z) for three sensors value are computed and then used in experiment with label attribute. The experiments are performed on one subject, ten subjects and all the subjects and compared in terms of accuracy, precision and recall. The results obtained from this study proves that KNN outperforms other machine learning methods in terms of accuracy, precision and recall. It is also concluded that personalization of data improves accuracy.

*Keywords:* ADLs, Datasets, Falls, IoT, Machine learning, Sum Vector Magnitude,

## 1. Introduction

The term IoT "The Internet of Things" is a network of devices connected via the internet. The concept is being applied in various fields such as pollution control, smart farming, smart cities, smart security, transportation and healthcare. The applications of IoT in healthcare are discussed in [1] and these applications are categorized into four domains: personal and healthcare, enterprise, utilities, and mobiles. The main reason to use IoT in healthcare is to improve the quality of life, reduce the cost of care, improve the access to care, provide healthcare tips to millions of people to live a healthy life and provide medical healthcare services to the patients. Initially, IoT was proposed for object identification (using RFID technology) and virtual representation [2]. Later it was used to monitor personalized healthcare. IoT based healthcare systems connect available resources to monitor, diagnose and predict health problems via the Internet [3]. The use of big data analytics in the IoT environment solves the problem of voluminous data storage [4]. Some anomalies in healthcare systems are discussed in [5].

Wearable devices are becoming popular for collecting physiological features of patient on regular basis which can be transferred to some storage facility such as cloud or databases (Thingspeak, Dropbox, MongoDB, IBM Bluemix, SQLite, MySQL, Hadoop etc.)  and analyzed to predict potential health issues. The patient's medical data can be used by doctors to provide recommendations. If a patient is in a serious condition, the ambulance facility may be provided and caretakers, nurses and family members be informed via messages or alarm generation. The data are sent to the cloud using communication technologies like Bluetooth, Wi-Fi, ZigBee, RFID, wireless LAN, USB, infrared, 6LowPAN and 5G mobile Network [6]. Some wireless technologies and cloud-based platforms for IoT are given in [7].

Advancements in healthcare provide quality of life specially to elderly population. In [8] the author provides information that the total population of elderly above 65 age is 962 million and it will extend up to two billion by 2050. This age group mostly suffer from mental, emotional, psychological and health issues. Chronic disease, Alzheimer's disease, Seizure disease, Diabetes, Dementia, Heart Attack are common in aged people. Person suffering from these types of diseases experience falls or unintentional injury performing Activities of Daily Livings (ADLs). The usage of medicines in routine may also impact their quality of life [9] and physical ability [10]. Accurately identifying ADLs and falls in elderly people using wearables can be a boon for people who live alone. Few studies on binary and multiclass classification of ADLs and falls are available [11, 12, 13, 14, 15, 16, 17, 18, 19, 20, 21, 22].

There are datasets created by researchers using wearables which are available for research community to evaluate the proposed systems/ frameworks. This study is an attempt to provide an overview of all the publicly available datasets which have been used for ADLs and falls. A study has also been conducted to evaluated Machine Learning methods (Logistic Regression, Linear discriminant analysis, K-Nearest Neighbor, Decision Tree and Naive Bayes) for detecting ADLs and falls. SisFall dataset has been used for conducting the study. It is observed that KNN works better than other techniques for individual or multiple subjects. As the dataset is generalized, the accuracy of the system decreases by more than 3.23% using KNN without SVM but with SVM it is decreased by 13.19 %. In this paper, an overview of all the datasets is given in Section 2. Experimental Evaluation of Machine Learning methods on SisFall dataset is given in Section 3. Results and Discussion are given in Section 4 with conclusion in Section 5.

## 2. Datasets

The publicly available datasets have been used in various studies. These datasets are related to falls, daily life activities, heart disease, maternal health and kidney disease. Datasets such as Smartwatch [13], Notch [13], Farseeing [12, 13, 34], MobiAct [18, 22], Usc-Had Database [12], Localization Data for Person Activity (LDPA) Dataset [12], The German Aerospace Center (Dlr) Human Activity Recognition Datasets [12], KTH dataset [11], SisFall [20,19,23], SmartFall Dataset [24], MobiFall [20,22], UR Fall dataset [21,16] and the UP-Fall dataset [21] are related to falls and ADLs. The datasets such as UCI machine learning repository, DaLiAc Dataset [25], mHealth Dataset [25], FSP Dataset [25], SBHAR Dataset [25], UbiqLog [26], CrowdSignals [26], ExtraSensory dataset [26], RFID Dataset [27], Smartphone Dataset [27], MobiAct RealWorld(HAR) [22], UMA Fall[ 22], Shoaib PA [22], Shoaib SA [22], tFall [22], UCI HAR [22], UCI HAPT [22], WISDM [22, 28], UniMiB SHAR [22,29], DMPSBFD [22] are related to daily living activities. Some datasets like BRFSS [30], PAMAP2 dataset [31] are related to chronic diseases. And other datasets such as the CASAS dataset (Tulum, Cairo, Milan) [32], KARD dataset [33], CAD-60 dataset [33], WISDM V.1.1 [28], WISDM V.2.0 [28], SKODA [28,19] are all IoT based publicly available datasets listed in Table 1. The description of IoT based publicly available datasets are given below:

1. **Smartwatch Dataset [13]:** The data are collected using MS Band2 sensor in the form of a smartwatch worn at the wrist. In this dataset, data are obtained from seven persons, aged between 21-55 years, height ranging between 5-6.5ft and weight ranging between 100lbs-230lbs. While performing falls and ADLs the process is repeated ten times by each subject.
2. **Notch Dataset [13]:** The data are collected using MS Band2 using multiple sensors mounted at the wrist of a person. The data is obtained from seven subjects aged between 20-35 years, height ranging between 5-6ft and weight ranging between 100lbs-200lbs while performing four types of falls and seven ADLs.
3. **Farseeing Dataset [13,12,34]:** The Farseeing stands for Fall Repository for the design of Smart and sElf- adaptive Environments prolonging Independent living. In this dataset, more than 300 real world fall events are recorded. The data are collected using ActivePal or McRobert Dynaport Mini mode sensors. This device was attached using a custom elastic belt at the waist. The data are captured from the subjects above 65 years old performing twenty-three types of falls at a sample rate of 20-100Hz.
4. **MobiAct Dataset[18,22]:** The data are collected using Samsung Galaxy S3 smartphone (accelerometer, gyroscope, and orientation sensors) with the LSM 330 DLC inertial module, fiftyseven subjects(forty-two men and fifteen women) aged between 20 to 47 years, the weight ranging between 50 kg to 120 kg and the height ranging between 160 cm to 189 cm. The subjects perform nine different types of ADLs and four falls with more than 2500 trials. These are back sitting chair, forward lying, front knee lying and sideward lying. Nine ADLs are car step-out, sit chair, standing, stairs down, walking, jumping, stairs up, car step-in and jogging.

5. **USC-HAD Database[12]:** In the University of Southern California-Human Activity Dataset (USCHAD) dataset, fourteen subjects(seven males and seven females) are used for data acquisition who were aged between 21 to 49 years, height ranging between 160-185cm and weight ranging between 43-80kg by performing twelve activities: sleeping, walking forward, sitting, walking left, walking right, running forward, walking upstairs, elevator up, walking downstairs, jumping, standing, elevator down and each subject performed five trails for each activity. A single MotionNode attached to a miniature laptop and a mobile phone pouch worn at the front right hip are used to collect data from the users.

6. **Localization Data for Person Activity (LDPA) Dataset [12]:** In this dataset, data are collected from five people while performing eleven activities like walking, falling, lying down, lying, sitting down, sitting, standing up from lying, sitting on the ground, standing up from sitting on the ground, standing up from sitting, that are repeated five times by each person while wearing four sensors (belt, ankle left, chest, and ankle right). This dataset consists of 164680 instances and eight attributes.

7. **The German Aerospace Center (DLR) HAR Dataset [12]:** For collecting the data, the IMU (3 axis accelerometer, 3 axis gyroscope, and a 3-axis magnetometer) are used. The data acquired from seventeen subjects aged between 23 and 50 years performed eleven ADLs and repeated many times by each person.

8. **KTH Dataset [11]:** In this dataset six different actions are performed by twenty-five subjects using a static camera in form of videos. These actions are walking, hand-clapping, running, boxing, hand waving, and jogging in four different scenarios i.e., outdoors s1, outdoors with scale variation s2, outdoors with different clothes s3 and indoors s4. In [11], the KTH dataset has been used to detect motion and evaluate the performance of the proposed model.

9. **SisFall Dataset [20,19,23]:** In this dataset, the data are collected from thirty-eight volunteers aged between 19-75 years. All the subjects, both old and young, performed nineteen types of ADLs and fifteen types of falls using Kinect MKL25Z128VLK4 microcontroller, an analog device: a Freescale MMA8451Q accelerometer, ADXL345 accelerometer and an ITG3200 gyroscope attached with the belt.

10. **SmartFall Dataset [24]:** In this dataset, the data are collected from 14 healthy subjects, aged between 21 to 60 years, weight ranging between 45-104kg and height between 152 cm to 198 cm. The data are collected using a wearable smartwatch embedded with an accelerometer sensor. The dataset is binary in nature. It contains labelled data having labels of "Fall" and "No Fall". This dataset consists of total 183,806 instances.

| Reference | Dataset | Subjects | Type of data | Sensors used |
| --- | --- | --- | --- | --- |
| [13] | Smartwatch | 7 Sub | Falls, ADLs | MS Band2 |
| [13] | Notch | 7 Sub | 7 ADLs, 4 Falls | MS Band2 |
| [13,12,34] | Farseeing | Subjects >65 age | 23 Falls | ActivePAL3, McRobert Dynaport minimode |
| [18,22] | MobiAct | 57 Sub (42M,15F) | 9 ADLs, 4 Falls | Smartphone |
| [12] | Usc-HAD | 14 Sub (7M,7F) | 12 ADLs | Single MotionNode, Miniature laptop |
| [12] | LDPA | 5 Sub | 11 ADLs | Wearing four tags (left, right ankle, belt, chest) |
| [12] | The German Aerospace Center (Dlr) | 16 Sub | 7 ADLs | Inertial sensor |
| [11] | KTH | 25 Sub | 6 ADLs | Static camera |
| [20,19,23] | SisFall | 38 Sub | 19 ADL, 15 Fall | 2 Accelerometer, Gyroscope |
| [24] | SmartFall | 14 Sub | ADLs and Falls | Smartwatch |
| [20,22] | MobiFall | 11 Sub (6M,5F) | ADLs and Falls | Smartphone |
| [21,16] | UR Fall | 5 Sub | 5 ADLs, 4 Falls | Microsoft Kinect cameras, Accelerometer |
| [21] | UP Fall | 17 Sub (9M,8F) | 6 ADLs, 5 Falls | Wearable, Ambient sensors, Vision devices |
| [25] | DaLiAc | 19 Sub (8F,11M) | 3 ADLs | 4 SHIMMER sensors |
| [25] | mHealth | 10 Sub | 12 ADLs | 4 Sensors |
| [25] | FSP | 10 Sub | 7 Activities | 5 Smartphones |
| [25] | SBHAR | 30 Sub | 6 ADLs | Smartphone |
| [26] | UbiqLog | 35 Sub | - | Smartphone, Smartwatch |
| [26] | CrowdSignals | 2 Sub | 8 Activities | Smartphone, Smartwatch |
| [26] | ExtraSensory | 60(34F, 26M) | 51 Activities | Smartphone, Smartwatch |
| [27] | RFID | >1000 | ADLs | RFID sensor |
| [27] | Smartphone | 30 Sub | 7 Activities | Smartphone |
| [22,18] | MobiAct RealWorld (HAR) | 54 Sub | 9 ADLs, 4 Falls | Smartphone |
| [22] | UMA Fall | 19 Sub | ADL, 3 Falls | 4 Bluetooth sensors motes, Smartphone |
| [22] | Shoaib PA, Shoaib SA | 4,10 Sub | ADLs | - |
| [22] | tFall | 10 Sub | ADLs, 4 Falls | Accelerometer |
| [22] | UCI HAR, UCI HAPT | 30 Sub, 30 Sub | 6 ADLs | Smartphone |
| [22,28] | WISDM V.1.1 and V.2.0 | 29 Sub | 6 ADLs | Smartphone |
| [22,29] | UniMiB SHAR | 30 Sub (6M,24F) | 9 ADLs, 8 Falls | Smartphone |
| [22] | DMPSBFD | 5 Sub | ADLs, Falls | Smartphone |
| [31] | PAMAP2 | 9 Sub | 18 Activities | 3 IMUs, Heart rate monitor |
| [32] | CASAS | 2 Sub | ADLs | - |
| [33] | KARD | 10 Sub | 18 Activities | Kinect sensor |
| [33] | CAD-60 | 4 Sub | 12 Activities | MS Kinect sensor |
| [28,19] | SKODA | 1 Sub | Gestures | 20 Accelerometer |

Table 1. Publicly available wearables-based Datasets.

11. **MobiFall Dataset[20,22]:** The data are recorded from twenty-four persons(seventeen males, seven females) aged between 22-47 years, weight ranging between 50-103 kg, height ranging between 1.60-1.89 m from inertial sensors(3D accelerometer and gyroscope) using Samsung Galaxy S3 smartphone by performing nine ADLs: standing, walking, jogging, jumping, stairs up, stairs down, sit chair, car step-in, car step-out and four falls: forward lying, front knees-lying, sideward lying and back sitting-chair.

12. **UR Fall Dataset [21,16]:** The data are collected using Microsoft Kinect cameras and accelerometer sensor. In this dataset, data are obtained from five subjects, ages over 26 years. While performing

    four types of falls: forwards while seated, forwards while walking, lateral fall while seated, lateral fall while walking and five ADLs: lying on the floor, lying on the sofa, sitting down, crouching down, picking up an object from the floor. A total of seventy (40ADLs /30 falls) samples are collected in this dataset.

13. **UP Fall Dataset [21]:** The data are collected using wearable sensors, ambient sensors and vision devices. In this dataset, data are obtained from seventeen (9 male and 8 female) subjects, aged between 18-24 years, height ranging between 1.57- 1.75m and weight ranging between 53-99kg while performing five types of falls: falling forward using hands, falling forward using knees, falling backward, falling sitting in an empty chair and falling sideward and six ADLs: walking, standing, picking up an object, sitting, jumping and laying.

14. **DaLiAc Database [25]:** DaLiAc(Daily Life Activities) database contains data from nineteen subjects in which eight are females and eleven are males. Each subject performed thirteen activities of daily life. The actions are sitting, lying, standing, washing dishes, vacuuming, sweeping, walking outside, ascending stairs, descending stairs, treadmill running, bicycling (50 watts), bicycling (100 watts) and rope jumping. Four SHIMMER sensors (3-axial accelerometer and 3-axial gyroscope) are used to record data from subjects with a 204.8Hz sampling rate and for segmentation 5s window length with a 50% overlapping window is used. Sensors are placed on the left ankle, right ankle, right hip and chest of the participants, and data are stored on the SD card.

15. **mHealth Dataset [25]:** The mobile Health dataset consists of twelve activities performed by ten subjects using three sensors (Shimmer wearable sensor) which are placed on the chest, right wrist, and left ankle for detection of the motion and vital signs of the body of subjects. The data are captured at the sampling rate of 50Hz, with 4s of window size with an overlapping. The activities selected for data collection are standing still, sitting and relaxing, lying down, walking, climbing stairs, waist bending forward, the frontal elevation of arms, knees bending(crouching), cycling, jogging, running, jump front and jump back.

16. **FSP Dataset [25]:** Five Smart Phone (FSP) dataset is captured using five smartphones at different body positions. Right jeans pocket, left jeans pocket and belt position towards the right leg positions are selected for daily life activities detection and the right upper arm position is selected for jogging, the right wrist of participants is used for a smartwatch. The seven activities performed by participants are walking, jogging, walking upstairs, downstairs, biking, etc. and the experiments are carried out indoor in university buildings, except biking. Ten subjects are chosen for experimental data collection at a sampling rate of 50Hz with 2s window length and the overlapping of 1s.

17. **SBHAR Dataset [25]:** The Smartphone-Based Human Activity Recognition (SBHAR) dataset is recorded using Samsung Galaxy S II with inertial sensors (accelerometer and gyroscope) mounted on the waist of thirty subjects with age ranging between 19-48 years while doing six ADLs at a frequency of 50Hz with 2.56s window length and overlapping of 1.28s.

18. **UbiqLog Dataset [26]:** In this dataset, data are collected using smartphones from the thirty-eight participants performing daily life activities for two months. It contains 9782222 instances. UbiqLog dataset was the first publicly available dataset based on smartphone lifelogging.

19. **CrowdSignals Dataset [26]:** The data are captured from more than thirty volunteers for thirty days from android users across the United States using smartphones and smartwatches.

20. **ExtraSensory Dataset [26]:** In this dataset, data are recorded using smartphones and smartwatches from thirty subjects (thirty-four females, twenty-six males) while doing daily life activities, age ranging between 18-42 years, height ranging between 145-188 cm and weight ranging between 50-93kg. Thirty-four iPhone users (iPhone gen: 4, 4S, 5, 5S, 5C, 6 and 6S) and twenty-six android users (Samsung, Nexus, HTC, Moto G, LG, Motorola, One Plus One, Sony), with more than 300k labeled examples describing posture and movement of the body and each subject is identified with UUID (universally unique identifier) with 1-minute interval. Smartphones with inbuilt sensors (accelerometer, gyroscope, magnetometer, watch accelerometer), location services, audio, watch compass, phone state indicators, and other additional sensors are used to acquire data.

21. **RFID Dataset [27]:** This dataset consists of various ADLs (lying on a bed, sitting or transitioning from bed to upright position, sitting on a chair and ambulating). The Data collected using the RFID sensor measure 3D acceleration and RSSI (received signal strength indicator) placed over the sternum.

22. **Smartphone Dataset [27]:** In this dataset, data are captured using in-built sensors (accelerometer and gyroscope) in smartphones placed inside the front trouser pocket of the subject. The data are collected from thirty subjects aged between 22-79 years performing seven activities at the rate of 50Hz. Activities such as standing, sitting, lying, walking, walking upstairs, walking downstairs are performed by each subject for 60seconds.

23. **MobiAct RealWorld (HAR) Dataset [18,22]:** The data of 54 participants while performing four falls and nine ADLs with more than 2500 trials are recorded using a smartphone Samsung Galaxy S3(accelerometer, gyroscope, and orientation sensors) and a mobile phone is kept in the trousers' pocket of the subject for capturing falls and ADLs. Four types of falls: forward lying, front knees lying, sideward lying, back sitting-chair and nine ADLs which are selected for the experiment are standing, walking, jogging, jumping, stairs up, stairs down, sit chair, car step-in and car step-out.

24. **UMA Fall Dataset [22]:** This dataset consists of falls and ADLs performed by nineteen subjects aged between 18-55 years, height ranging between 155-195 cm, weight ranging between 50-93 kg. The data are collected in the form of videos using accelerometers, gyroscopes, and magnetometer within smartphones and four external IMUs. In this dataset eight types of ADLs that are selected for the experiment are: body bending (squatting), climbing downstairs, climbing upstairs, hopping, light jogging, lying down (and getting up) on(from) a bed, sitting down (and up) on(from) a chair, walking at a normal pace and three falls included in the dataset are: backward, forward and lateral fall.

25. **Shoaib PA, Shoaib SA Dataset [22]:** This dataset is recorded using four Samsung Galaxy S2 smartphones with inertial sensors (accelerometers, gyroscopes, and magnetometer) attached to the trousers' pocket of the participants, arm, wrist, and the belt of seven subjects performing seven types of activities. These activities are: walking, running, standing, sitting, cycling, walking upstairs and downstairs and the sampling rate used for activities is 50Hz.

26. **tFall Dataset [22]:** In this dataset, data are captured from ten subjects (seven males and three females) aged between 22 to 32 years, two smartphones used by the subjects in daily life for ADLs and falls. In this dataset eight types of falls are performed by the subjects.
27. **UCI HAR Dataset [22]:** Human Activity Recognition dataset is collected from thirty volunteers aged between 19-48 years using a Samsung Galaxy S II smartphone (accelerometer and gyroscope) mounted at the waist of subjects while performing six activities standing, sitting, laying down, walking, walking downstairs and upstairs. Each activity is performed twice by each subject during the experiment, at the sampling rate of 50 Hz.
28. **UCI HAPT Dataset [22]:** UCI HAPT dataset is updated versions of UCI HAR in which the data are recorded from thirty subjects of age ranges between 19-48 while performing ADLs using original raw inertial signals from the smartphone sensors, instead of the signals pre-processed into windows which were provided in version 1. The activities are: three static postures (standing, sitting, lying) and three dynamic activities (walking, walking downstairs and walking upstairs).
29. **WISDM Dataset [22,28]:** WISDM (WIreless Sensor Mining) dataset is collected using android smartphones (3-axial accelerometer) containing 1098207 instances, six attributes and twenty-nine users. The users are selected to perform six activities of daily life (walking, jogging, upstairs, downstairs, sitting and standing).
30. **UniMiB SHAR Dataset [22,29]:** Using smartphone (Samsung Galaxy Nexus I9250), data are collected from thirty subjects, their age ranging between 18-60 years, height ranging between 160190cm, weight ranging between 50-82kg doing nine ADLs, eight falls and containing 11771 samples at the sampling rate of 50Hz. Nine ADLs are standing up from laying, lying down from standing, standing up from sitting, running, sitting down, going downstairs, going upstairs, walking, jumping and 8 types of falls are falling backward-sitting-chair, generic falling backward, falling with protection strategies, generic falling forward, falling rightward, falling leftward, hitting an obstacle in the fall etc. that are performed during the time of data collection.
31. **DMPSBFD Dataset [22]:** DMPSBFD stands for Dataset for Mobile Phone Sensing Based Fall Detection. Data are collected using five smartphones with (accelerometer, gyroscope) by the five subjects while performing one ADLS and seven falls at the sampling frequency of 5Hz and repeated five times by each subject. Five smartphones that were used during experiments are Samsung Galaxy S4, Samsung Galaxy S5 Mini, Sony Z3, Google Nexus 4 and Google Nexus 5.
32. **BRFSS Dataset [30]:** Behavioral Risk Factor Surveillance System dataset captures the data related to health behavior for chronic conditions and disorders. It is a collaborative project of the Centres for Disease Control and Prevention (CDC) of states and territories of the United States. In this dataset, data was collected through the telephonic interview from adults to measure risk factor behavior in household living. It collects data from fifty states, three US territories and districts of Columbia. More than 400,000 persons participate every year.
33. **PAMAP2 Dataset [31]:** Physical Activity Monitoring dataset captures the data of activity recognition from nine subjects performing 18 activities at the sampling rate of 100Hz using three IMUs and heart rate monitor placed over the wrist, chest and ankle of the body. Data files contain 52 attributes, 54 columns with activity labels: lying, sitting, computer work, standing, walking, running, nordic walking, cycling, watching TV, ascending stairs, descending stairs, car driving, folding laundry, ironing and vacuum cleaning.
34. **CASAS Dataset (Tulum, Cairo, Milan) [32]:** Centre for Advanced Studies in Adaptive Systems (CASAS) dataset is a project for creating real smart homes using ambient motion sensors and installed toolkit in thirty-two homes for the generation of datasets. In this dataset five ADLs: make a phone call, wash hands, cook, eat and clean are performed by the participants. CASAS contains 64 datasets such as Tulum, Cairo, Milan, Kyoto, Tokyo, Paris, Aruba, Scone, HH101HH130, Shib010-Shib020 and Shibsdf. In [35] Tulum, Cairo, Milan datasets are used. In the Tulum dataset, two participants are selected to capture daily life activities, two participants with one pet are selected to record daily life activities in the Cairo dataset and one participant with one pet are used for daily living in Milan dataset.

35. **KARD Dataset [33]:** In this dataset, data are recorded in a controlled environment using a Kinect sensor placed at a distance of two-three meter from the subjects. Eighteen activities are performed by ten young (nine males, one female) subjects, aged between 20-30 years, height ranging between 150-185 cm. These activities are repeated three times by each subject that can be divided into ten gestures: two-hand wave, horizontal arm wave, high throw, high arm wave, draw X, side kick, draw tick, forward kick, hand clap, bend, and eight actions (take an umbrella, catch cap, walk, toss the paper, phone call, sit down drink, and stand up). The dataset is composed of RGB and depth frames (30 fps) with resolution 640*480.

36. **CAD-60 Dataset [22]:** This dataset composed of RGB, depth frames and skeleton data using Kinect sensor in an indoor environment. Twelve activities are performed by four subjects (two males and two females) in five different environments (kitchen, bathroom, living room, office and bedroom). These activities are wearing contact lens, rinsing the mouth, talking on the phone, brushing teeth, drinking water, opening pill container, cooking-stirring, cooking-chopping, talking on the couch, working on the computer, relaxing on the couch and writing on a whiteboard.

37. **WISDM V.1.1 Dataset [28,19]:** In this dataset, data are obtained from twenty-nine users doing six different activities using acceleration sensors at the sampling rate of 20Hz. A total of 91515 samples are collected in this dataset.

38. **WISDM V.2.0 Dataset [28,19]:** The data are captured from thirty-six users doing six different activities using acceleration sensors at the sampling rate of 20Hz. A total of 248653 samples are collected in this dataset.

39. **SKODA Dataset [28,19]:** The dataset is recorded using acceleration sensors at four different body positions of the user performing ten activities at the sampling rate of 96Hz and the data are collected from one user. In this dataset, the total samples collected are 22000.

## 3. Experimental Evaluation

SisFall dataset is selected for evaluating the performance of machine learning methods to detect ADLs and Falls accurately. The experimentation is performed using Numpy, Matplotlib, Scikit-learn and Pandas libraries of python. The techniques are compared with or without calculating magnitudes of all the three axes. Total six experiments are performed. For first, third and fifth experiment all the 9 attributes of dataset are considered binary classification (0-ADLs, 1-Fall). In second, fourth and sixth experiment, Sum Vector Magnitude value for all the three sensors (2 Accelerometers and one gyroscope) using the formula given in (1) is calculated and used.

**Sum Vector Magnitude:** Sum Vector Magnitude (SVM) is used to summarize the vector. SVM value is determined by summing up squares of all the three axes i.e. x, y, z axis and calculating the square root of this value.

$$|A| = \sqrt{(x_1^2 + y_1^2 + z_1^2)} \qquad (1)$$

### 3.1 Dataset Description:

- **SisFall dataset:** This dataset consists of total 4510 text files which contain data from 38 subjects. Further 38 participants are divided into two groups: 15 elderly people between 60 and 75 years of age and 23 adults between 19 and 30 years of age. performing 19 types of ADLS and 15 Falls. Total 34 activities are performed by every participant. These activities and falls data are acquired using self-developed device consisting of Kinets MKL25Z128VLK4 microcontroller, analog devices, SD card for storage, 1000 mA/h generic battery, two accelerometer models ADXL345 which has a resolution of 13 bits and a range of −16g to +16g, and Freescale MA8451Q which has a resolution of 14 bits and a range of −8g to +8g and one gyroscope ITG3200 which has 16 bits and a range of -2000°/s to + 2000°/s. These devices are placed on participants waist. Pair of 3-axis accelerometers are used for acceleration data and 3-axis gyroscope for rotation data. The structure of dataset folder and files are given in figure 1 and figure 2.

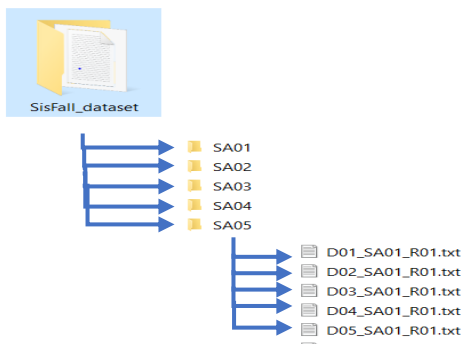
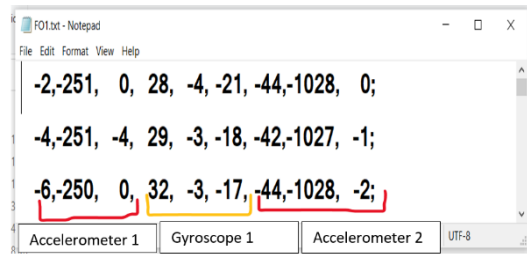

Figure 1. Folders and files format of SisFall dataset.     Figure 2. Attributes used for ADLS and Falls in SisFall dataset.

ADLs and Falls performed during data collection are Walking slowly, Walking quickly; Jogging slowly; Jogging quickly; Walking upstairs and downstairs slowly; Walking upstairs and downstairs quickly; Slowly sit in a half height chair, wait a moment, and up slowly; Quickly sit in a half height chair; wait a moment, and up quickly; Stumble while walking; Gently jump without falling; Fall forward while walking caused by a slip; Fall backward while walking caused by a slip; Lateral fall while walking caused by a slip; Fall forward while walking caused by a trip; Fall forward while jogging caused by a trip; Fall forward when trying to get up; Fall backward when trying to sit down etc. as shown in figure 3.

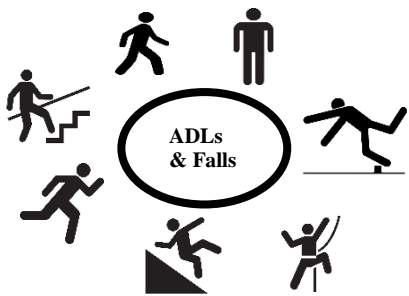

Figure 3. ADLs and Falls performed during data collection.

### 3.2 Machine Learning Methods

Five Machine learning: Logistic Regression, Linear Discriminant Analysis, K-Nearest Neighbor, Decision Tree and Naive Bayes algorithms are applied and compared. Best classifier is selected using k-fold cross validation method. The dataset is divided into training and testing with the ratio of 80 and 20 respectively for all the six experiments given in Table 2. Machine learning methods used in this study are discussed below:

- **Logistic Regression (LR):** LR is a supervised learning algorithm used for statistical regression analysis. It is used for classification to measure the association between one dependent variable and 'one or more' independent variable. It is applied for both binary and multiclass classification. In [9], multivariate logistic regression is applied to analysis the risk of falls. In [30], the author concludes that LR with ANN is an effective tool for the prediction of hypertension among patients suffering from chronic diseases. The model achieved 71.96% accuracy.
- **Linear Discriminant Analysis (LDA):** LDA is also known as Discriminant Function Analysis. It is supervised learning based technique used for dimensionality reduction. It converts the higher dimensional space into lower space by removing the dependent and redundant features before classification. It is applied in healthcare, face recognition, customer identification etc. As discussed in [9], for classification, LDA has good generalization capability with less computational cost. In [14], Subspace Linear Discriminant Analysis is used for binary and multiclass classification.
- **K-Nearest Neighbor:** KNN is a non-parametric method and supervised machine learning algorithm, used for solving both classification and regression problems. This algorithm is used when the dataset is labeled, noise-free and small. The algorithm provides a solution to the problem of identifying similar objects. Euclidean distance, Mahalanobis distance or Hamming distance are applied to find the new data points. The process of choosing the value of k is known as parameter tunning (odd value of k is always selected to avoid the confusion between two classes) [36]. The main advantages of this algorithm are simple, easy implementation, versatile and there is no need to build a model. If the number of parameters increases, this algorithm works slowly. In [37], both KNN and Decision Tree algorithms are applied for fall detection, the data are collected using accelerometer and gyroscope sensors connected with Arduino and transferred the data on MangoDB for further processing. KNN algorithm performs better than the Decision Tree in detecting falls with 98.75% and 90.59% accuracies respectively.
- **Decision Tree (DT):** DT is a supervised machine learning algorithm. It is a non-parametric algorithm that is widely used for classification and prediction. In this algorithm, the decisions are represented in the form of tree and flowchart like structure using the divide and conquer method. Each branch of the tree represents a possible decision. The output is in the form of leaf nodes. It is used in decision making [37] and can handle both numerical and categorical data. In [38], accelerometer-based movement and fall detection system uses decision tree algorithm to detect ten movements of the body. This algorithm succeeds in defining positions of the body with 81.48% accuracy.
- **Naive Bayes:** It requires fewer number of data points to be trained, deals with high dimensional data points, fast, used in real-time, handles both discrete and continuous data, and supports scalability. It is applied in spam filtering, text categorization, news classification, weather prediction, and automatic medical diagnosis [36]. In [13], Naive Bayes classifies falls in less than 0.3 seconds.

Table 2. Six experiments performed on different subjects using Machine Learning methods.

| S.No | Experiments | LR (acc %) | LDA (acc %) | KNN (acc %) | DT (acc %) | NB (acc %) | MODEL selected (Using K fold cross validation) |
|---|---|---|---|---|---|---|---|
| 1 | 1 subject | 99.09 | 99.05 | 96.53 | 98.76 | 71.77 | 96.53% |
| 2 | 1 subject with SVM | 62.62 | 59.69 | 82.92 | 79.26 | 56.81 | 83.01% |
| 3 | 10 subjects | 77.52 | 77.48 | 94.57 | 89.29 | 76.01 | 94.85% |
| 4 | 10 subjects with SVM | 70.42 | 70.42 | 73.42 | 67.01 | 70.42 | 73.42% |
| 5 | All subjects | 76.92 | 77.01 | 92.98 | 87.19 | 76.48 | 93.30% |
| 6 | All subjects with SVM | 66.91 | 66.53 | 69.81 | 63.75 | 65.98 | 69.82% |

### 3.3 Evaluation Measures

The measures used for evaluation are: Accuracy, Precision, Recall and F1-score. The Precision, Recall and F1-score for ADLs and Falls for different subjects are given separately in Table 3.

$$\textbf{Accuracy} = TP + TN/(TP + TN + FP + FN)/100$$

$$\textbf{F1-score} = 2*recall*precision\ (recall + precision)$$

$$\textbf{Precision} = TP/(TP+ FP)$$

$$\textbf{Recall} = TP/TP+FN$$

Table 3. Precision, Recall, F1-score of ADLs and Falls on different subjects.

| ACTIVITIES | PERFORMANCE MEASURES | 1 SUBJECT | 10 SUBJECTS | ALL SUBJECTS |
|---|---|---|---|---|
| ADLs=0 | *Precision* | 96% | 96% | 94% |
| | *Recall* | 98% | 97% | 96% |
| | *F1-Score* | 97% | 96% | 95% |
| Falls=1 | *Precision* | 97% | 93% | 92% |
| | *Recall* | 95% | 89% | 88% |
| | *F1-Score* | 96% | 94% | 90% |

### 4. Results and Discussion

The accuracies achieved for all the five methods over 1 subject, 10 subjects and All subjects are shown in Table 2. The results prove that KNN model classify ADLs and Falls more accurately as compared to other four ML methods. In this experimentation, we analogize KNN with SVM and without SVM and it is observed that the accuracy is decreased using SVM as given in Table 4. Here comparisons are made with different subjects and results prove that when the number of subjects is increasing, the accuracy is decreasing. If we compare results of KNN without SVM using 1 subject and 10 subjects the accuracy is decreased by 1.68% and between 10 Subjects and all the subjects it is decreased by 1.55%. And if we compare KNN with SVM in term of subjects the accuracy is decreased by 3.6% to 9.59%. From this study, it is concluded that as we move towards generalization it effects the results. Personalization improves accuracy.

Table 4. Comparison of KNN with or without SVM between different subjects in terms of accuracy.

| Models | 1 SUBJECT | 10 SUBJECTS | ALL SUBJECTS |
|---|---|---|---|
| KNN (without magnitude) | 96.53% | 94.85% | 93.30% |
| KNN (with magnitude) | 83.01% | 73.42% | 69.82% |

### 5. Conclusion

The elderly people mostly suffer from health problems. They mostly experience falls too. The use of a health monitoring system in daily life to detect falls using features related to activities of daily living can provide better health monitoring decisions and quality of life. The main contribution of this paper is to provide a review of the datasets created using wearables to monitor activities of daily living and falls. Using SisFall dataset performance of five machine learning methods are evaluated and it is observed that KNN is a good classifies for ADLs and Fall prediction. It achieves 93.30% accuracy for all subjects. In future, the plan is to implement multiclass classification on wearable dataset for fall detection. Physiological features can be explored in combination with ADLs for better system development.

# References

[1] Kulkarni, Alok and Sampada Sathe (2014) "Healthcare applications of the Internet of Things: A Review." International Journal of Computer Science and Information Technologies: Vol. 5 (5), 6229-6232.

[2] Qi, Jun, Po Yang, Geyong Min, Oliver Amft, Feng Dong and Lida Xu. (2017) "Advanced internet of things for personalised healthcare systems: A survey." Pervasive and Mobile Computing: 1574-1192. https://doi.org/10.1016/j.pmcj.2017.06.018.

[3] YIN, Yuehong, Yan Zeng, Xing Chen and Yuanjie Fan. (2016) "The internet of things in healthcare: An overview." Journal of Industrial Information Integration: 3-13. https://doi.org/10.1016/j.jii.2016.03.004.

[4] Firouzi, Farshad, Amir M. Rahmani, K. Mankodiya, M. Badaroglu G.V. Merrett, P. Wong and Bahar Farahani. (2018) "Internet-of-Things and big data for smarter healthcare: From device to architecture, applications and analytics." Future Generation Computer Systems: 583-586. https://doi.org/10.1016/j.future.2017.09.016.

[5] Fahim, Muhammad and Alberto Sillitti. (2019) "Anomaly Detection, Analysis and Prediction Techniques in IoT Environment: A Systematic Literature Review." IEEE Access: Institute of Electrical and Electronics Engineers Inc VOLUME 7-2169-3536. https://doi.org/10.1109/ACCESS.2019.2921912.

[6] Chin, Jeannette, Alin Tisan, Victor Callaghan and David Chik. (2021) "Smart-Object-Based Reasoning System for Indoor Acoustic Profiling of Elderly Inhabitants." Electronics: vol 10,1433.

[7] Fernandez, Felipe and George C. Pallis. (2014) "Opportunities and challenges of the Internet of Things for healthcare." National Conference on Wireless Mobile Communication and Healthcare - Transforming healthcare through innovations in mobile and wireless technologies. https://doi.org/10.4108/icst.mobihealth.2014.257276.

[8] Thakur, Nirmalya, Y. Han, Chia. (2022) "A Simplistic and Cost-Effective Design for Real-World Development of an Ambient Assisted Living System for Fall Detection and Indoor Localization: Proof-of-Concept." Information. 13, 363. https://doi.org/10.3390/info13080363.

[9] Oya, Nozomu, Ayani, Nobutaka, Kuwahara, Akiko, Kitaoka, Riki, Omichi, Chie, Sakuma, Mio, Morimoto, Takeshi and Narumoto, Jin. (2022) "Over Half of Falls Were Associated with Psychotropic Medication Use in Four Nursing Homes in Japan: A Retrospective Cohort Study." Int. J. Environ. Res. Public Health. 19, 3123 (2022). https://doi.org/10.3390/ijerph19053123.

[10] Choi, Ahnryul, Kim, Tae Hyong, Yuhai, Oleksandr and Mun, Joung Hwan. (2022) "Deep Learning-Based Near-Fall Detection Algorithm for Fall Risk Monitoring System Using a Single Inertial Measurement Unit." IEEE Trans. Neural Syst. Rehabil. Eng. Vol.30, 2385–2394 (2022). https://doi.org/10.1109/TNSRE.2022.3199068.

[11] Hsieh, Yi-Zeng and Yu-Lin Jeng. (2017) "Development of Home Intelligent Fall Detection IoT System Based on Feedback Optical Flow Convolutional Neural Network." IEEE Access: 6, 6048–6057. https://doi.org/10.1109/ACCESS.2017.2771389.

[12] Liu, Zhi, Yankun Cao, Lizhen Cui, Jiahua Song, and Guangzhe Zhao. (2018) "A Benchmark Database and Baseline Evaluation for Fall Detection Based on Wearable Sensors for the Internet of Medical Things Platform." IEEE Access: vol 6, 51286–51296. https://doi.org/10.1109/ACCESS.2018.2869833.

[13] R. Mauldin, Taylor, Marc E. Canby, Vangelis Metsis, Anne H. H. Ngu and Coralys Cubero Rivera. (2018) "Smartfall: A smartwatch-based fall detection system using deep learning." Sensors (Switzerland): vol 18, 1–19. https://doi.org/10.3390/s18103363.

[14] Kavuncuoglu, Eehan, Uzunhisarcıklı, Esma, Barshan, Billur, Ozdemir, Ahmet Turan (2022) "Investigating the Performance of Wearable Motion Sensors on recognizing falls and daily activities via machine learning." Digit. Signal Process. Vol 126, 103365. https://doi.org/10.1016/j.dsp.2021.103365.

[15] Thakur, Nirmalya, and Chia Y. Han. (2021) "A Study of Fall Detection in Assisted Living: Identifying and Improving the Optimal Machine Learning Method." Journal of Sensor and Actuator Networks: vol 10, 39.

[16] Chen, Yangsen, Rongxi Du, Kaitao Luo and Yuheng Xiao. (2021) "Fall detection system based on real-time pose estimation and SVM." Proceedings - IEEE 2nd International Conference on Big Data, Artificial Intelligence and Internet of Things Engineering: vol 978-1-6654-1540-8/21. https://doi.org/10.1109/ICBAIE52039.2021.939006.

[17] Garcia-Moreno, Francisco, Bermudez-Edo, Maria and Garrido, Jose Luis (2020) "A Microservices e-Health System for Ecological Frailty Assessment Using Wearables." Sensors Vol 1–23. https://doi:10.3390/s20123427

[18] Ajerla, Dharmitha, Sazia Mahfuz and Farhana Zulkernine. (2019) "A real-time patient monitoring framework for fall detection." Wireless Communications and Mobile Computing: vol 2019, Article ID 9507938. https://doi.org/10.1155/2019/9507938.

[19] Yacchirema, Diana, Jara Suarez de Puga, Carlos Palau and Manuel Esteve. (2019) "Fall detection system for elderly people using IoT and ensemble machine learning algorithm." Personal and Ubiquitous Computing. https://doi.org/10.1007/s00779-018-01196-8.

[20] He, Jian, Zihao Zhang, Xiaoyi Wang, and Shengqi Yang. (2019) "A low power fall sensing technology based on fd-cnn." IEEE Sensors Journal: Vol. 19, NO. 13, 5110–5118. https://doi.org/10.1109/JSEN.2019.2903482.

[21] M. Galvao, Yves, Janderson Ferreira, Vinicius A. Albuquerque, Pablo Barros and Bruno J.T. Fernandes. (2021) "A multimodal approach using deep learning for fall detection." Expert Systems with Applications: vol 168,114226. https://doi.org/10.1016/j.eswa.2020.114226.

[22] Micucci, Daniela, Marco Mobilio and Paolo Napoletano. (2017) "UniMiB SHAR: A dataset for human activity recognition using acceleration data from smartphones." Applied Science: 7, 1101. https://doi.org/10.3390/app7101101.

[23] Jin Park, Yu, Seol Young Jung, Tae Yong Son and Soon Ju Kang. (2021) "Self-Organizing IoT Device-Based Smart Diagnosing Assistance System for Activities of Daily Living." Sensors (Switzerland): vol 21, 785. https://doi.org/10.3390/s21030785.

[24] Farsi, Mohammed. (2021) "Application of ensemble RNN deep neural network to the fall detection through IoT environment." Alexandria Engineering Journal: vol 60, 199–211. https://doi.org/10.1016/j.aej.2020.06.056.

[25] Zdravevski, Eftim, Petre Lameski, Vladimir Trajkovik and Andrea Kulakov. (2017) "Improving Activity Recognition Accuracy in Ambient-Assisted Living Systems by Automated Feature Engineering." IEEE Access: 5, 5262–5280. https://doi.org/10.1109/ACCESS.2017.2684913.

[26] Vitabile, Salvatore, Michal Marks and Dragan Stojanovic. (2019) "Medical Data Processing and Analysis for Remote Health and Activities Monitoring." Medical Data Processing and Analysis: 186–220. https://doi.org/10.1007/978-3-030-16272-6_7.

[27] Awais, Muhammad, Mohsin Raza, Kamran Ali, Zulfiqar Ali, Muhammad Irfan, Omer Chughtai, Imran Khan, Sunghwan Kim and Masood Ur Rehman. (2019) "An internet of things based bed-egress alerting paradigm using wearable sensors in elderly care environment." Sensors (Switzerland): 19, 1–17. https://doi.org/10.3390/s19112498.

[28] Zheng, Xiaochen, Meiqing Wang and Joaquín Ordieres-Mere. (2018) "Comparison of data preprocessing approaches for applying deep learning to human activity recognition in the context of industry 4.0." Sensors (Switzerland): 18, 2146. https://doi.org/10.3390/s18072146.

[29] Al-kababji, Ayman, Abbes Amira, Faycal Bensaali, Abdulah Jarouf, Lisan Shidqi, Hamza Djelouat. (2021) "Biomedical Signal Processing and Control An IoT-based framework for remote fall monitoring." Biomedical Signal Processing and Control: vol 67, 102532. https://doi.org/10.1016/j.bspc.2021.102532.

[30] Wang, Aiguo, Ning An, Yu Xia, Lian Li and Guilin Chen. (2014) "A Logistic Regression and Artificial Neural Network-based Approach for Chronic Disease Prediction: A Case Study of Hypertension." IEEE International Conference on Internet of Things, Green Computing and Communications, and Cyber-Physical-Social Computing: 978-1-4799-5967-9.

[31] Khowaja, Sunder Ali, Aria Ghora Prabono, Feri Setiawan, Bernardo Nugroho Yahya and Seok-Lyong Lee. (2018) "Contextual activity based Healthcare Internet of Things, Services, and People (HIoTSP): An architectural framework for healthcare monitoring using wearable sensors." Computer Networks:145, 190–206. https://doi.org/10.1016/j.comnet.2018.09.003.

[32] Al Zamil, Mohammed GH, Majdi Rawashdeh, Samer Samarah, M. Shamim Hossain, Awny Alnusair and SK MD Mizanur Rahman. (2017) "An Annotation Technique for In-Home Smart Monitoring Environments." IEEE Access: 6, 1471–1479. https://doi.org/10.1109/ACCESS.2017.2779158.

[33] Saini, Rajkumar, Pradeep Kumar, Barjinder Kaur, Partha Pratim Roy, Debi Prosad Dogra and K. C. Santosh. (2019) "Kinect sensor-based interaction monitoring system using the BLSTM neural network in healthcare." International Journal of Machine Learning and Cybernetics: 10, 2529–2540. https://doi.org/10.1007/s13042-018-0887-5.

[34] Yin, Cunyi, Jing Chen, Xiren Miao, Hao Jiang and Deying Chen. (2021) "Device-free human activity recognition with low-resolution infrared array sensor using long short-term memory neural network." Sensors: vol 21. https://doi.org/10.3390/s21103551.

[35] Wickramasinghe, Asanga, Damith C. Ranasinghe, Christophe Fumeaux, Keith D. Hill and Renuka Visvanathan. (2017) "Sequence Learning with Passive RFID Sensors for Real-Time Bed-Egress Recognition in Older People." IEEE Journal of Biomedical and Health Informatics: 21, 917–929. https://doi.org/10.1109/JBHI.2016.2576285.

[36] Saeid Mahdavinejad, Mohammad, Rezvan, Mohammadreza and Barekatain Mohammadamin (2018) "Machine learning for internet of things data analysis: a survey." Digit. Commun. Networks. Vol. 4, 161–175. https://doi.org/10.1016/j.dcan.2017.10.002.

[37] Kumar Bhoi, Sourav, Sanjaya Kumar Panda, Bivash Patra and Bijaya Pradhan. (2018) "FallDS-IoT: A Fall Detection System for Elderly Healthcare Based on IoT Data Analytics." In: Proceedings - 2018 International Conference on Information Technology, ICIT 2018. pp.: 155–160. Institute of Electrical and Electronics Engineers Inc. https://doi.org/10.1109/ICIT.2018.00041.

[38] Ichwana, Dody, M. Arief and Nefy Puteri. (2018) "Movements Monitoring and Falling Detection Systems for Transient Ischemic Attack Patients Using Accelerometer Based on Internet of Things." In Proceedings -2018 International Conference on Information Technology Systems and Innovation (ICITSI) Bandung - Padang. October 22-25, 2018 ISBN: 978-1-5386-5692-1.